\documentclass[runningheads]{llncs}
\usepackage[T1]{fontenc}
\usepackage{graphicx}
\usepackage{float}
\usepackage{booktabs}
\usepackage{amsmath,amssymb}
\usepackage{xcolor}
\usepackage{hyperref}
\usepackage[capitalise,noabbrev]{cleveref}
\hypersetup{hidelinks, pdfauthor={Souraj Adhikary, Negar Chabi, Andre Mastmeyer}, pdftitle={Bound-Aware Per-Organ Recall Risk Control}, pdfsubject={Conformal risk control for multi-organ CT segmentation under clinical domain shift}, pdfkeywords={conformal risk control, distribution shift, recall guarantees, multi-organ segmentation}}
\usepackage{placeins} 

\begin{document}

\title{Bound-Aware Per-Organ Recall Risk Control for Multi-Organ CT Segmentation under Clinical Domain Shift}
\titlerunning{Bound-Aware Per-Organ Recall Risk Control}

\author{Souraj Adhikary\inst{1} \and
Negar Chabi\inst{1} \and
Andre Mastmeyer\inst{1}}

\authorrunning{S.~Adhikary, N.~Chabi, A.~Mastmeyer}
\institute{Department of Engineering Sciences, Jade Hochschule Wilhelmshaven, Germany\\
\email{souraj.adhikary@student.jade-hs.de}\\
\email{\{negar.chabi,andre.mastmeyer\}@jade-hs.de}}

\maketitle

\begin{figure}[H]
\centering
\includegraphics[width=0.82\textwidth]{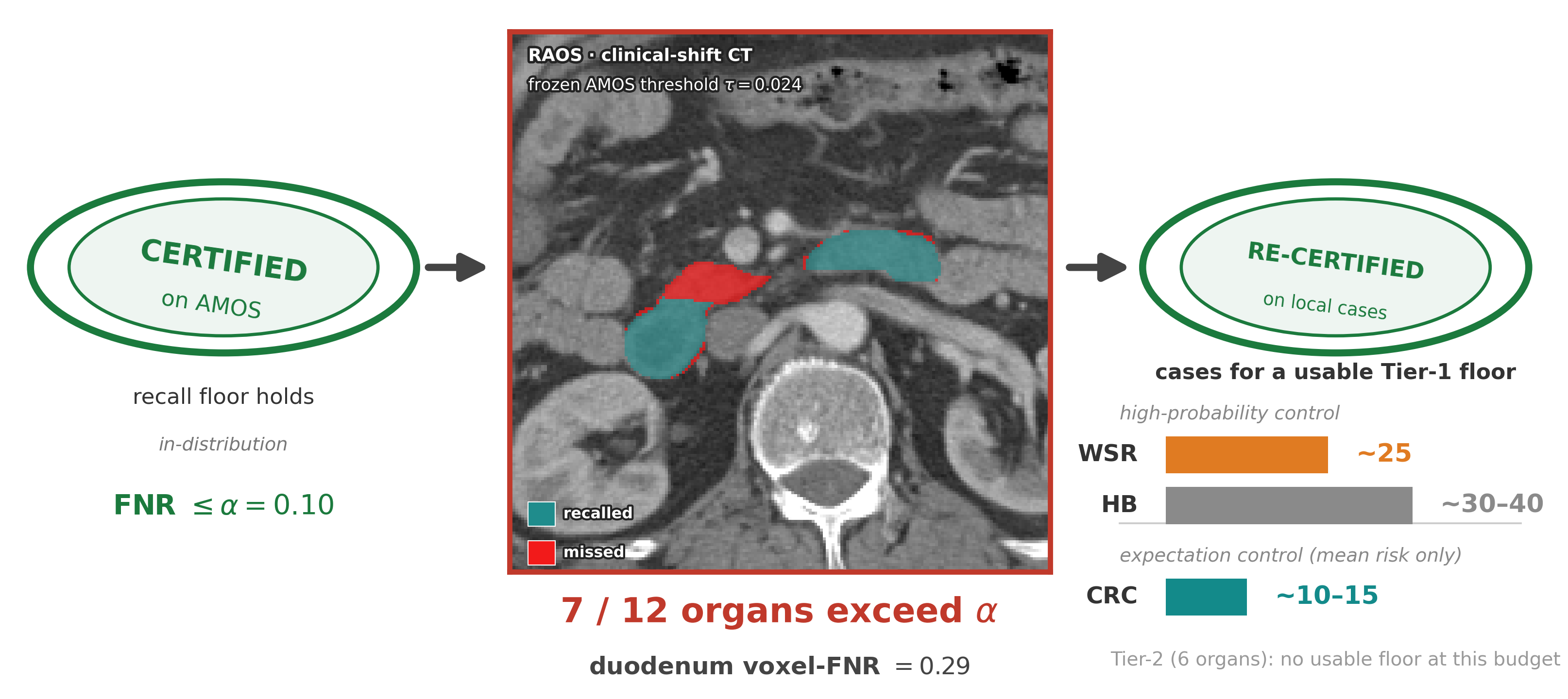}
\caption{\textbf{Frozen AMOS thresholds often fail on RAOS.} \emph{Centre:} one RAOS case at the duodenum threshold $\tau{=}0.024$; red marks missed ground-truth voxels, teal recalled ones (case FNR $0.29$, cohort mean $0.30$). With pooled AMOS out-of-fold calibration, $7/12$ organs exceed $\alpha{=}0.10$ on RAOS; only three of the other five have a non-vacuous threshold. Here, re-certification varies with the bound and budget (\Cref{sec:e6}).}
\label{fig:teaser}
\end{figure}

\begin{abstract}
Distribution-free risk control adds organ-specific recall guarantees to frozen segmentation. We calibrate per-organ thresholds for an AMOS-trained nnU-Net, audit transfer to RAOS, and estimate local re-certification cost using case-level voxel false-negative rate (FNR). The AMOS control passes, but $7/12$ organs exceed $\alpha{=}0.10$ after transfer; smaller calibration sets can mask exceedances with conservative or vacuous thresholds. Risk-Controlling Prediction Sets (RCPS) give high-probability control of population-mean risk, whereas Conformal Risk Control (CRC) gives weaker expectation control. Both require exchangeability; fixed and global thresholds give no per-organ guarantee. The Waudby--Smith--Ramdas (WSR) betting bound re-certifies six Tier-1 organs with 25 local cases, versus 30--40 for Hoeffding--Bentkus (HB). CRC needs 10--15 but has a heavier individual-case tail. No Tier-2 organ meets our illustrative precision criterion with 25 cases.
\end{abstract}

\section{Introduction}

A high mean Dice score cannot show whether every organ-at-risk is recalled well enough for radiotherapy contouring or surgical planning~\cite{isensee2021,wasserthal2023totalseg}. Softmax scores can also be poorly calibrated~\cite{mehrtash2020calibration}. Building on conformal prediction~\cite{vovk2005,lei2018distfree,angelopoulos2023gentle}, distribution-free risk control sets post-hoc thresholds with finite-sample guarantees on a chosen risk. We study Risk-Controlling Prediction Sets (RCPS)~\cite{bates2021rcps} and Conformal Risk Control (CRC)~\cite{angelopoulos2024crc}. Applied to each organ's voxel false-negative rate (FNR $=1-$recall), these thresholds define a \emph{recall floor}: the guaranteed quantity is the population-mean fraction of ground-truth organ voxels missed, not every patient's loss.

These guarantees require exchangeable calibration and test cases. Differences in disease, scanner, or imaging protocol can break this assumption, so a research-cohort threshold may fail in clinical deployment. We ask: \emph{how many labelled local cases are needed to re-certify the guarantee, and how much over-segmentation does the threshold produce?}

We study an AMOS-trained~\cite{ji2022} nnU-Net on RAOS~\cite{raos2024}, an extension of WORD~\cite{word2022} that labels organ presence and absence across a graded surgical-shift ladder. We contribute: (1)~a per-organ recall-risk formulation and in-distribution positive control; (2)~a transfer audit that separates the effects of cohort shift and calibration size; (3)~evidence that fixed and global thresholds provide no per-organ guarantee; and (4)~a local re-certification analysis that keeps high-probability and expectation-level control separate.

\section{Related Work}

\textbf{Distribution-free risk control.} RCPS~\cite{bates2021rcps} calibrates a threshold so a bounded risk is controlled with high probability via a concentration bound; CRC~\cite{angelopoulos2024crc} controls it in expectation; Learn-then-Test~\cite{angelopoulos2021ltt} handles multiplicity across a grid. Our third rule is the variance-adaptive WSR betting bound~\cite{waudbysmith2024betting}.
\textbf{Conformal under shift.} Covariate-shift weighting~\cite{tibshirani2019shift} and beyond-exchangeability bounds~\cite{barber2023beyond} relax the usual assumptions. They do not directly solve our setting because we cannot assume that $P(Y\,{\mid}\,X)$ is unchanged; correcting only the input distribution $P(X)$ may be insufficient. We instead re-certify on labelled local data, which may already exist where clinics contour these organs for radiotherapy. Prior work has likewise shown that conformal guarantees can degrade by subgroup under shift~\cite{mehrtens2023pitfalls}.

\textbf{Conformal/risk control for segmentation.} Recent work builds conformal prediction sets for segmentation masks~\cite{mossina2025morphconformal}, predicts Dice ranges for quality control~\cite{wundram2024conformalqc}, and applies CRC to FNR/FDR-style risks~\cite{dai2025fdrcrc}.

\textbf{The two closest works.} sem-CRC~\cite{teneggi2025semcrc} controls an expectation-level semantic risk in CT within one distribution. We instead choose a softmax threshold for each organ, compare high-probability RCPS with expectation-level CRC, and test both after exchangeability fails. COMPASS~\cite{cheung2026compass} uses representation-based importance weights to build conformal intervals for mask-derived quantities such as organ area under covariate shift. Its interval summarizes a fixed mask; our threshold changes the mask to meet a recall-risk target. Its weighting targets $P(X)$ and may not cover changes in $P(Y\,{\mid}\,X)$. Representation-based weighting could still complement local re-certification, but we do not test that combination.

\textbf{Segmentation quality control and hallucination.} RAOS~\cite{raos2024} shows that AMOS-trained networks can fail on shifted clinical scans, and frozen nnU-Net models can fail silently out of distribution~\cite{gonzalez2021nnunetfail}. HALOS~\cite{rickmann2023halos} retrains a model to suppress hallucinated organs after resection. We do not retrain; we certify recall only for organs known to be present and study whether that guarantee transfers. Standard Dice and HD95 do not express this guarantee~\cite{metrics2024}.

\section{Method}

\subsection{Per-organ recall risk}
For organ $k$ in a case where that organ is present, we measure the fraction of its ground-truth voxels missed at threshold $\tau$:
\begin{equation}
L_k(\tau) = 1 - \frac{|\{v\in G_k: p_k(v)\ge\tau\}|}{|G_k|},
\qquad
R_k(\tau) = \mathbb{E}_{\mathrm{case}}\!\left[L_k(\tau)\right],
\label{eq:loss}
\end{equation}
Here $p_k(v)$ is the softmax score for organ $k$ at voxel $v$; measuring only inside $G_k$ prevents background dilution. $L_k$ is one case's loss, whereas $R_k$ is the population mean across cases. Every guarantee below constrains $R_k$, not each patient's $L_k$. Cases, rather than voxels, are the exchangeable units. Lowering $\tau$ can only recall more voxels, so $L_k\in[0,1]$ is non-decreasing in $\tau$. This floor applies only to organs known to be present and does not control absent-organ hallucinations.

Unlike softmax argmax, which assigns one class per voxel, we threshold each organ channel separately. Lower thresholds miss fewer organ voxels but add predicted-positive voxels, so we report both effects.

\subsection{Calibrating the recall floor}
For each organ, we seek the highest threshold that still certifies a population-mean missed-voxel rate no greater than $\alpha$; a higher threshold usually gives fewer false positives. Under RCPS~\cite{bates2021rcps}, we select the largest $\lambda_k$ whose $(1-\delta)$ upper confidence bound (UCB) on risk is at most $\alpha$:
$\lambda_k = \sup\{\tau:\ \mathrm{UCB}_{1-\delta}(R_k(\tau))\le\alpha\}$. Because risk is monotone in $\tau$, acceptable thresholds form a prefix of the grid and need no grid-wide multiplicity correction~\cite{bates2021rcps}. The selected threshold satisfies
\begin{equation}
\mathbb{P}_{\mathrm{cal}}\!\left(R_k(\lambda_k)\le\alpha\right)\ \ge\ 1-\delta,
\label{eq:guarantee}
\end{equation}
where $\lambda_k$ depends on the random calibration sample and $\mathbb{P}_{\mathrm{cal}}$ is taken over that sample.

\textbf{What the guarantee means.} Equation~\eqref{eq:guarantee} says that at least a $1-\delta$ fraction of calibration samples produce a threshold whose population-mean future loss is at most $\alpha$. It does not promise $L_k\le\alpha$ for every patient. CRC instead gives $\mathbb{E}[L_k(\lambda_k)]\le\alpha$~\cite{angelopoulos2024crc}, where the expectation includes both the calibration sample and future case and has no $\delta$. CRC may need fewer cases because it controls only this overall expectation. Its guarantee is not interchangeable with high-probability RCPS control.

\textbf{The three selection rules.} We compare two RCPS bounds, HB and WSR, with CRC. Each rule selects the largest grid point $\lambda_k$ satisfying its condition. Here $\hat{R}$ is mean FNR across calibration cases, $L_i$ is case $i$'s loss, and $n$ is the number of calibration cases:
\begin{center}
\small
\begin{tabular}{@{}l@{\quad}l@{}}
HB~\cite{bates2021rcps} & $\min\{e^{-n h_1(\hat{R}\wedge\alpha,\,\alpha)},\ e\,\mathbb{P}[\mathrm{Bin}(n,\alpha)\le\lceil n\hat{R}\rceil]\}\le\delta$ \\[1pt]
WSR~\cite{waudbysmith2024betting} & $\max_{t\le n}\prod_{i\le t}\left(1+\eta_i(\alpha-L_i)\right)\ge 1/\delta$ \\[1pt]
CRC~\cite{angelopoulos2024crc} & $(n\hat{R}+B)/(n{+}1)\le\alpha$ \\
\end{tabular}
\end{center}
HB is the standard RCPS bound: it is tight when observed risk is near zero but does not use observed variance. WSR is variance-adaptive and can be tighter when losses have low variance, as they often do for a recall floor. Here $h_1(a,b)=a\log(a/b)+(1{-}a)\log\big((1{-}a)/(1{-}b)\big)$ and $B{=}1$ bounds the loss. The $\wedge$ clamp prevents HB from certifying a threshold once $\hat{R}\ge\alpha$. WSR has one tuning choice, the predictable betting fraction $\eta_i=\sqrt{2\log(1/\delta)/(n\hat\sigma^2_{i-1})}$. The lagged variance $\hat\sigma^2_{i-1}$ makes $\eta_i$ depend only on $L_1,\dots,L_{i-1}$; we cap it at $1/(1-\alpha)$ so each factor remains non-negative. We tune no other hyperparameter.

\textbf{Marginal, not simultaneous, control.} We calibrate each organ independently. The guarantee is therefore marginal for one organ; it does not control the chance that at least one of 12 organs exceeds $\alpha$. For HB and WSR, replacing $\delta$ with $\delta/12$ gives simultaneous high-probability control by a union bound; we report its extra cost in \Cref{sec:e6}. CRC has no $\delta$ to divide, so simultaneous expectation control needs a different method that we do not study. Finite samples also limit which risks can be certified. With $\delta{=}0.10$, HB needs about 45 cases for $\alpha{=}0.05$ and 22 for $\alpha{=}0.10$, independent of the model; WSR needs 46/23, and CRC needs $\lceil1/\alpha-1\rceil=19/9$.
\subsection{Transfer and bound-aware recalibration}
\label{sec:method-recal}
We first freeze the AMOS-calibrated $\lambda_k$ and measure FNR on RAOS (the transfer audit). We then re-certify each organ on $n$ local RAOS cases, repeating the analysis over random splits and values of $n$. Because all recalibration data come from one RAOS site, these are optimistic single-site estimates.

A recall floor does not limit over-segmentation. We therefore report whole-volume precision $=\mathrm{TP}/\mathrm{PP}$ and false-positive ratio $\mathrm{FP}_k=\max(\mathrm{PP}-\mathrm{TP},0)/|G_k|$ for present organs, where $\mathrm{PP}$ is the predicted-positive count and $\mathrm{TP}=\mathrm{recall}\cdot|G_k|$. Once the sample is large enough, $\tau\to0$ can control FNR by predicting almost everything. We instead report the \emph{usable} budget: the smallest $n$ that controls held-out FNR---in at least a $1-\delta$ fraction of draws for HB and WSR, or in mean for CRC---and gives mean held-out precision $\ge0.5$. This precision cutoff is illustrative, not clinically validated. In deployment, the appropriate cutoff may differ by organ; we did not evaluate organ-specific cutoffs and leave them to future clinical validation.

\textbf{How we define the organ groups.} At the reference setting (WSR, $\alpha{=}0.10$, $n{=}25$), six organs meet both conditions: liver, spleen, both kidneys, stomach, and bladder. We call them \textbf{Tier-1}; the other six form \textbf{Tier-2}. These labels summarize this experiment and are not clinical categories.

All three methods use the same threshold search, with local RAOS cases replacing AMOS. HB and WSR differ only in the finite-sample bound applied to the same losses; this can change the selected threshold, false-positive cost, and required sample size. CRC also provides a different, weaker guarantee. We therefore report CRC separately and do not treat its case counts as equivalent to HB or WSR counts.

\section{Experimental Setup}
\textbf{Model and data.} We use a 3D full-resolution nnU-Net~\cite{isensee2021} trained on AMOS CT~\cite{ji2022}. AMOS and RAOS use the same nnU-Net plan and preprocessing, but their predictors differ. AMOS calibration uses out-of-fold validation softmax: each case is predicted by the single fold model that held it out (236 of 243 cases are usable; seven corrupted exports are skipped). This cross-fitted set is not the one fixed predictor assumed by the formal guarantee. The main RAOS analysis instead averages all five fold models. A threshold calibrated on one predictor has no formal guarantee for another, so we also run a matched fold~0 analysis (\Cref{sec:transfer}).

The shifted cohort contains all 163 RAOS~\cite{raos2024} cases: Set~1 has 67 oncology cases, Set~2 has 22 partial-excision cases, and Set~3 has 74 full-excision cases. All calibration and risk-control analyses use cached softmax outputs after training; we do not retrain the model.

\textbf{Shared label space.} We study only the 12 organs defined consistently in AMOS and RAOS. We exclude the merged \texttt{prostate\_uterus} class, which would create spurious failures, and the aorta and IVC, which occur only in AMOS. Sex and contrast are audit strata only. FNR is defined only when an organ is present, so each organ has fewer than 163 eligible cases and resection can reduce that number. From Set~1 to Set~3, for example, right-kidney presence falls from 67 to 63 cases and gallbladder presence from 67 to 34. The recall loss therefore excludes errors caused by organ absence or resection.

\textbf{Protocol.} We use $\alpha\in\{0.05,0.10\}$ and $\delta=0.10$; an AMOS calibration/test split of 150/86 cases repeated over 20 seeds; 1001 thresholds in steps of $0.001$; 2000 case-level bootstrap samples (seed 2026) for transfer-audit CIs; and 200 random re-certification draws per budget. The illustrative precision criterion is $0.5$.

\section{Results}

\subsection{In-distribution positive control}\label{sec:posctrl}
The empirical AMOS positive control passes. Over 20 calibration/test splits of the 236 cases, every organ has realized test FNR $\le\alpha$ in at least 95\% of splits; the target is $1-\delta=0.90$. Mean held-out FNR, pooled across organs and both $\alpha$ values, is $0.028$. This split-refit result differs from the in-sample $\hat{R}(\lambda_k)\!\approx\!0.06$ in \Cref{tab:thresholds}. CRC makes no $1-\delta$ claim, so a lower split-wise rate, such as $0.55$ for the stomach, is not a violation. Tier-2 organs (\Cref{sec:e6}) already select $\tau\!\approx\!0$ on AMOS, showing that their low attainable precision is not caused only by transfer.

Simpler thresholds provide no per-organ control even in distribution. A fixed $\tau{=}0.5$ violates the floor for 8/12 organs at $\alpha{=}0.05$ and 6/12 at $\alpha{=}0.10$. One global threshold violates it for 5--6/12 organs: it is too conservative for easy organs and insufficient for hard ones.

\subsection{The floor does not survive transfer}\label{sec:transfer}
\begin{figure}[t]
\centering
\includegraphics[width=0.82\textwidth]{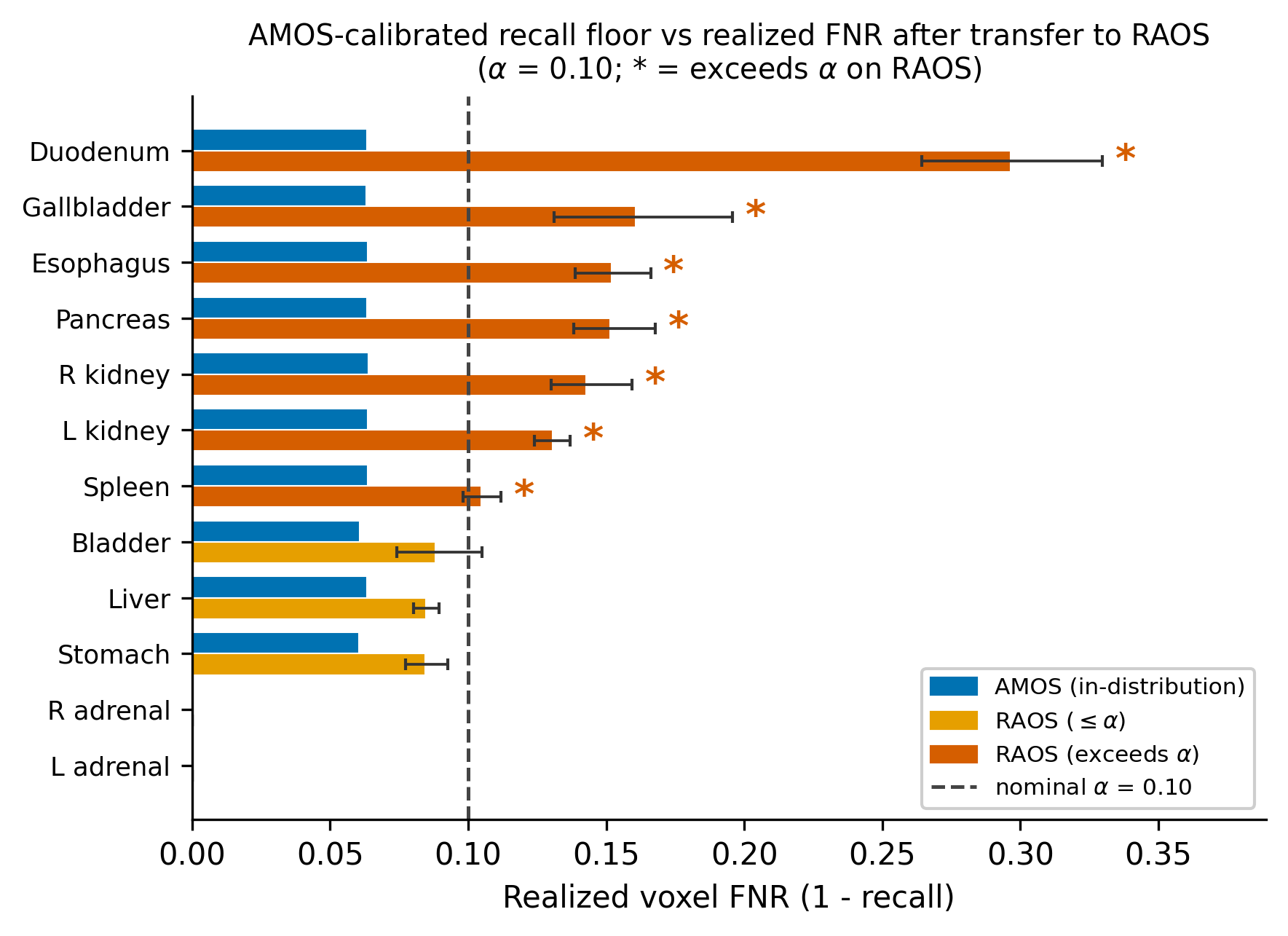}
\caption{Transfer audit ($\alpha{=}0.10$). Per-organ FNR for AMOS in-distribution vs.\ frozen-threshold transfer to RAOS, with 95\% CIs; highlighted organs exceed $\alpha$ on RAOS.}
\label{fig:transfer}
\end{figure}

When the frozen AMOS thresholds are applied to RAOS, cohort-mean FNR exceeds $\alpha$ for 6/12 organs at $\alpha{=}0.05$ and 7/12 at $\alpha{=}0.10$ (\Cref{fig:transfer,tab:thresholds}). Of the other five organs at $\alpha{=}0.10$, both adrenals have the vacuous predict-everything threshold $\lambda_k{=}0.000$. Only bladder, liver, and stomach remain below $\alpha$ with a non-vacuous threshold. This non-exceedance is an observed result, not proof that control transfers.

The change differs by organ. Duodenum FNR rises from $0.063$ to $0.297$, and right-kidney FNR from $0.064$ to $0.143$; both pairs have non-overlapping CIs. AMOS FNR is well below $\alpha$, so the higher RAOS values are not caused by an already loose in-distribution fit.

\textbf{The predictor mismatch has limited effect here.} Using fold~0 alone changes per-organ FNR by $0.002$ on average ($\le\!0.006$) and flips no exceedance verdicts ($7/12$ and $6/12$ in both arms). Fully matched model~0 calibration and deployment also agrees, giving $2/12$ against the ensemble's $2/12$ at the same 59-case budget. Matching the predictor does not restore exchangeability.

\textbf{Calibration size changes the exceedance count.} With 59 AMOS calibration cases over 500 draws, typical exceedances fall from $7/12$ to $2/12$ at $\alpha{=}0.10$ and from $6/12$ to $0/12$ at $\alpha{=}0.05$. This apparent improvement comes from wider margins lowering $\lambda_k$ and increasing predictions: the liver falls from $0.892$ to $0.614$ and from $0.525$ to $0.013$, respectively. At $\alpha{=}0.05$, 59 cases barely clear the 45-case floor and vacuous thresholds rise from $3/12$ to $8/12$. At $\alpha{=}0.10$, the right kidney, duodenum, pancreas, and left kidney still exceed in $72.6\%$, $71.6\%$, $23.2\%$, and $17.6\%$ of draws; each other organ stays below $16\%$.

\begin{table}[tb]
\centering
\setlength{\tabcolsep}{5pt}
\caption{Frozen AMOS thresholds $\lambda_k$ and realized voxel FNR at $\alpha{=}0.10$ on AMOS and the clinically shifted RAOS cohort, with 95\% CI. Boldface marks RAOS FNR exceeding $\alpha$; both adrenals' $\lambda_k{=}0.000$ is vacuous; $n_{\text{AMOS}}/n_{\text{RAOS}}$ are present-organ case counts.}
\label{tab:thresholds}
\begin{tabular}{lrrlrr}
\toprule
Organ & $\lambda_k$ & AMOS FNR & RAOS FNR (95\% CI) & $n_{\text{AMOS}}$ & $n_{\text{RAOS}}$ \\
\midrule
Bladder & 0.277 & 0.061 & 0.088 [0.074, 0.105] & 233 & 163 \\
Duodenum & 0.024 & 0.063 & \textbf{0.297 [0.264, 0.330]} & 236 & 163 \\
Esophagus & 0.050 & 0.064 & \textbf{0.152 [0.139, 0.166]} & 235 & 161 \\
Gallbladder & 0.051 & 0.063 & \textbf{0.161 [0.131, 0.196]} & 222 & 123 \\
L adrenal & 0.000 & 0.000 & 0.000 [0.000, 0.000] & 236 & 163 \\
L kidney & 0.690 & 0.064 & \textbf{0.130 [0.124, 0.137]} & 235 & 160 \\
Liver & 0.892 & 0.063 & 0.085 [0.080, 0.089] & 236 & 163 \\
Pancreas & 0.107 & 0.063 & \textbf{0.152 [0.138, 0.168]} & 236 & 163 \\
R adrenal & 0.000 & 0.000 & 0.000 [0.000, 0.000] & 236 & 156 \\
R kidney & 0.741 & 0.064 & \textbf{0.143 [0.130, 0.159]} & 234 & 152 \\
Spleen & 0.861 & 0.064 & \textbf{0.105 [0.098, 0.112]} & 235 & 163 \\
Stomach & 0.282 & 0.061 & 0.085 [0.077, 0.093] & 233 & 163 \\
\bottomrule
\end{tabular}
\end{table}

\begin{figure}[t]
\centering
\includegraphics[width=0.95\textwidth]{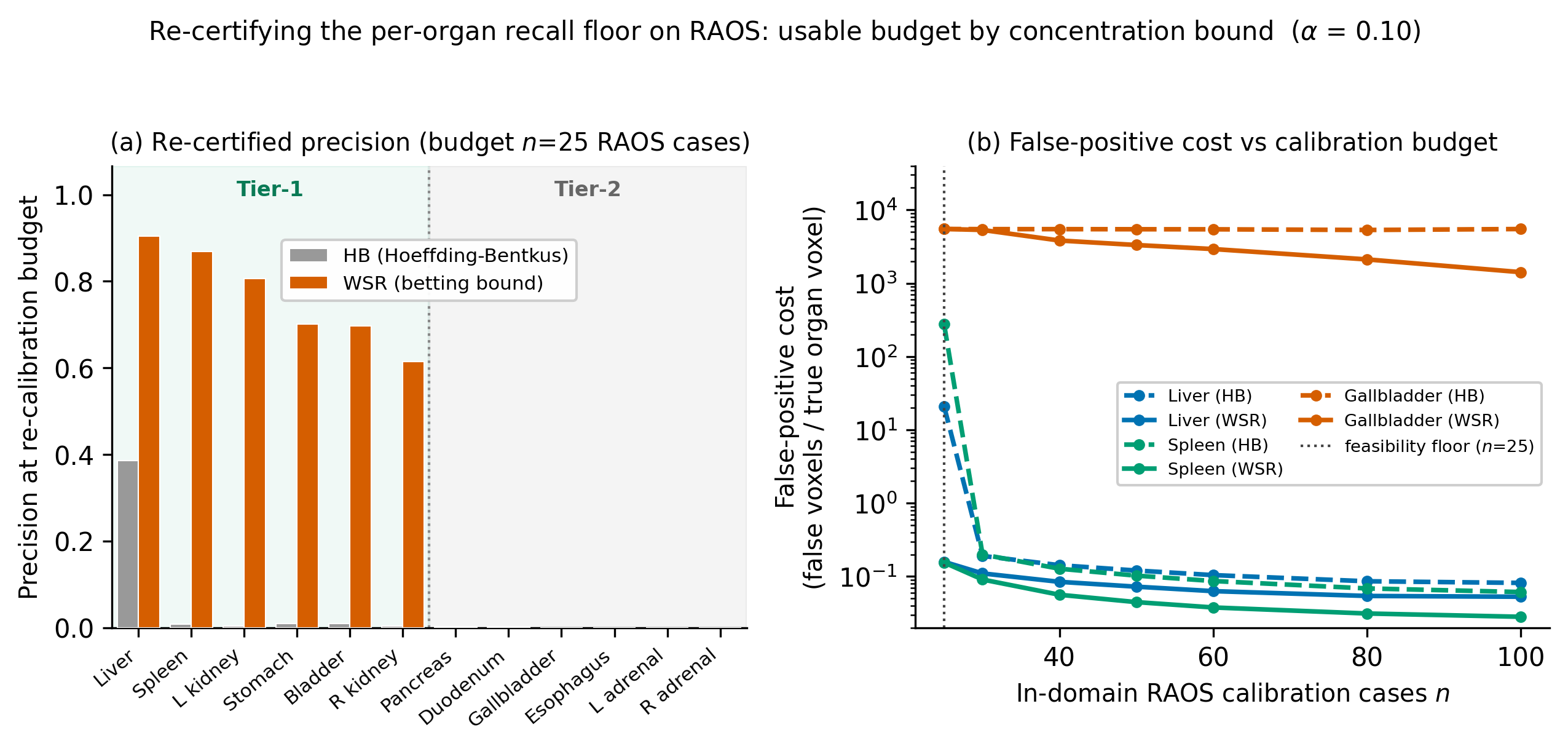}
\caption{Re-certification analysis ($\alpha{=}0.10$). (a)~Precision at a fixed 25-case budget, HB vs.\ WSR. (b)~False-positive cost vs.\ in-domain calibration cases.}
\label{fig:budget}
\end{figure}

\begin{table}[tb]
\centering
\small
\setlength{\tabcolsep}{9pt}
\caption{\textbf{Within high-probability control, WSR needs fewer cases than HB.} \emph{Left:} usable local RAOS budgets for the six Tier-1 organs; ranges span organs. A usable threshold controls held-out FNR and has precision ${\ge}0.5$; ``simult.'' uses Bonferroni $\delta/12$ over all 12 at $\alpha{=}0.10$. HB and WSR provide high-probability control; CRC provides expectation control, so their counts are not equivalent. \emph{Right:} held-out case loss at $n{=}25$, $\alpha{=}0.10$. Each column reports the worst statistic among organs with a non-vacuous threshold (CRC~9, WSR~6, HB~2); columns can refer to different organs. At $\alpha{=}0.05$, HB never certifies the right kidney within our grid.}
\label{tab:bounds}
\begin{tabular}{@{}lrrr@{\hspace{20pt}}rrrrr@{}}
\toprule
& \multicolumn{3}{c}{Usable budget $n$} & \multicolumn{5}{c}{Held-out loss, column-wise worst} \\
\cmidrule(lr){2-4}\cmidrule(lr){5-9}
Rule & $\alpha{=}.10$ & $\alpha{=}.05$ & simult. & mean & $p_{95}$ & $p_{99}$ & max & \%${>}\alpha$ \\
\midrule
\multicolumn{9}{l}{\emph{High-probability control:} $\mathbb{P}(R_k\le\alpha)\ge 1-\delta$} \\
WSR & 25 & 50--80 & 50--60 & .012 & .043 & .092 & .64 & 1.0 \\
HB & 30--40 & 60--100 & 60--100 & .002 & .010 & .020 & .29 & 0.0 \\
\addlinespace[3pt]
\multicolumn{9}{l}{\emph{Expectation control:} $\mathbb{E}[L_k]\le\alpha$} \\
CRC & 10--15 & 25 & --- & .066 & .175 & .922 & 1.00 & 17.4 \\
\bottomrule
\end{tabular}
\end{table}

\subsection{Within our grid, WSR needs fewer cases than HB}
\label{sec:e6}
At $\alpha{=}0.10$, HB and WSR need about 22--25 local RAOS cases before any threshold can be certified (\Cref{fig:budget,tab:bounds}); this reflects finite-sample limits. With the illustrative precision criterion, WSR re-certifies all Tier-1 organs with 25 cases and HB with 30--40---a $1.2$--$1.6\times$ range within high-probability control. CRC meets its separate expectation-level criterion with 10--15 cases; because this guarantee is weaker, its counts are not equivalent.

\textbf{At a fixed 25-case budget, WSR gives a usable floor and HB does not.} WSR attains precision $0.62$--$0.90$ (liver $0.90$, spleen $0.87$), whereas HB selects a near-pass-through threshold (liver $0.39$, $\le 0.01$ for the rest). The resulting false-positive cost differs by $4$--$1800\times$. That band is organ-dependent and its top end is one organ: right kidney $3.6\times$, liver $133\times$ ($21\!\to\!0.16$ FP voxels per true voxel), spleen $1789\times$. The re-certified WSR floor still controls recall: held-out RAOS FNR stays an order of magnitude below $\alpha$ (\Cref{tab:bounds}) and was $\le\alpha$ in all 200 draws for all six organs.

\textbf{Expectation control can hide poor individual cases.} Pooling held-out case losses over 200 draws at $n{=}25$ (\Cref{tab:bounds}, right), CRC keeps mean risk below $\alpha$, but up to $17.4\%$ of individual cases exceed $\alpha$. The 99th percentile is far above the mean, and the worst gallbladder and right-kidney cases approach a total miss. WSR at the same budget exceeds $\alpha$ on at most $1.0\%$ of cases; HB yields a non-degenerate threshold for only two organs. CRC's smaller case count therefore accompanies a much heavier individual-case tail. Its expectation guarantee is not equivalent to high-probability RCPS control.

\textbf{Simultaneous control over all 12 organs.} Replacing the marginal $\delta$ with a Bonferroni $\delta/12$ roughly doubles the usable budget (\Cref{tab:bounds}), and costs one organ: counting all 12, the number reaching the usability criterion anywhere within the ${\le}100$-case grid falls from 8 to 7 for WSR and from 7 to 6 for HB. Those totals exceed the six Tier-1 organs because pancreas and esophagus also qualify, but only at large budgets. At $\alpha{=}0.05$ a simultaneous statement is out of reach within the grid: HB certifies none of the 12 and WSR only the liver. Covering every organ at once is therefore materially more expensive than the per-organ budgets suggest.

\textbf{Tier-2 organs stay out of reach.} The six Tier-2 organs (pancreas, duodenum, gallbladder, esophagus, both adrenals) do not reach a usable floor at the 25-case reference budget. Pancreas and esophagus need larger calibration sets (WSR $\sim$40/$\sim$80 cases; CRC $\sim$20/$\sim$40), while gallbladder, duodenum and both adrenals never cross the precision criterion within our grid.

\textbf{The groups are stable across several precision cutoffs.} At the reference point, Tier-1 precision is at least $0.615$ and Tier-2 at most $0.0019$, so any cutoff in $(0.002,\,0.615]$, applied globally or per organ, preserves the groups. Budgets are stable from $0.25$ to $0.5$; at $0.75$, bladder, stomach, and right kidney move to Tier-2.

\begin{figure}[H]
\centering
\includegraphics[width=0.72\textwidth]{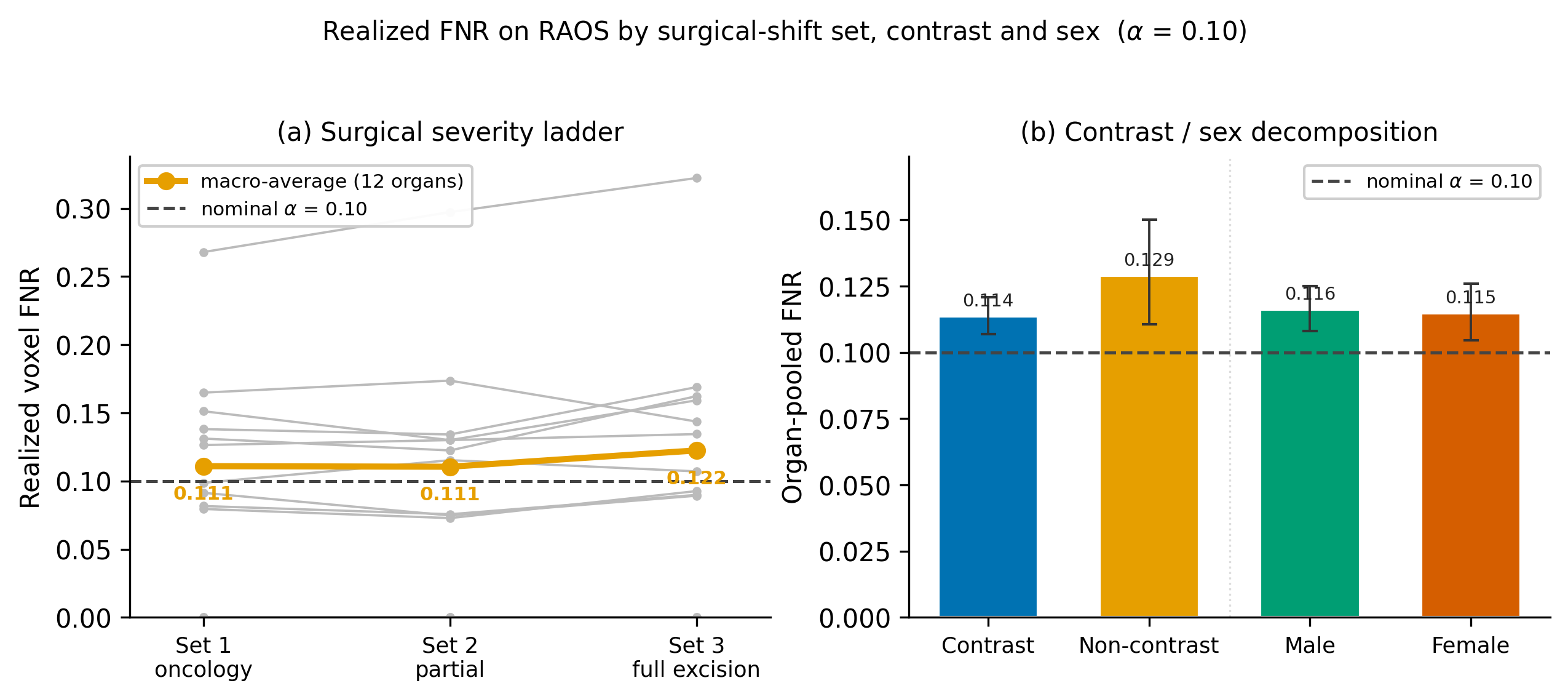}
\caption{(a)~Surgical-shift ladder is nearly flat. (b)~Contrast/sex decomposition with $\alpha$ and 95\% CIs.}
\label{fig:ladder}
\end{figure}

\textbf{Which shift axis is associated with the exceedances?} Macro-averaged FNR is nearly flat across Sets 1--3 ($0.111/0.111/0.122$; \Cref{fig:ladder}), so exceedances are not concentrated along the surgical gradient. Contrast and sex are likewise weak axes. Missing scanner and protocol descriptors preclude attribution to a specific mechanism.

\section{Discussion}
\textbf{What this adds beyond a shift warning.} An audit must report calibration size, vacuous thresholds, the deployed predictor, guarantee type, and false-positive cost. Here the RCPS bound changed the observed local sample size needed for a useful threshold. Our illustrative results motivate prospective testing of a Tier-1 workflow with $\sim$25 local cases and manual review for Tier-2 organs; they do not validate a clinical referral rule.

\textbf{Limitations.} We study one model and one source--target pair; generalization requires other architectures, modalities, and cohorts. We do not compare re-certification with retraining, domain adaptation, or other uncertainty methods. The main calibration and deployment predictors differ, and the matched check covers one fold of one model. Recalibration uses one site, requires labels of unmeasured cost, and yields budget estimates without confidence intervals.

The recall floor is not a full safety case: control is marginal per organ (and expectation-level for CRC), covers only known-present organs, and excludes hallucination and resection error. Independently thresholded masks may overlap, so whole-volume precision does not establish label-map usability. A per-organ ROI could reduce false-positive counts; its effect on the Tier-1/Tier-2 grouping remains untested.

\section{Conclusion}
AMOS-calibrated per-organ recall floors did not reliably transfer to RAOS. WSR met our high-probability criterion with fewer local cases than HB; CRC needed fewer under its weaker expectation guarantee but had a heavier individual-case tail. Tier-2 organs required larger sets or never met the illustrative precision criterion. The practical insight is that, once shift broke the guarantee, the choice of concentration bound---not the frozen model---set the local annotation budget needed to restore it in our audit, making bound selection a reportable design decision in deployment-time re-certification.

\FloatBarrier 

\begin{credits}
\subsubsection{\ackname}
This work was funded by the Deutsche Forschungsgemeinschaft (DFG), project number MA 6791/1-1.

\subsubsection{\discintname}
The authors have no competing interests to declare that are relevant to the content of this article.
\end{credits}

\bibliographystyle{splncs04}

\end{document}